\pdfoutput=1

\documentclass[11pt]{article}

\usepackage{acl}

\usepackage{times}
\usepackage{latexsym}

\usepackage[T1]{fontenc}

\usepackage[utf8]{inputenc}

\usepackage{microtype}

\title{Counterfactual Explanations for Natural Language Interfaces}

\author{George Tolkachev \\
  University of Pennsylvania \\
  \texttt{georgeto@seas.upenn.edu} \\\And
  Stephen Mell \\
  University of Pennsylvania \\
  \texttt{sm1@seas.upenn.edu} \\\AND
  Steve Zdancewic \\
  University of Pennsylvania \\
  \texttt{stevez@seas.upenn.edu} \\\And
  Osbert Bastani \\
  University of Pennsylvania \\
  \texttt{obastani@seas.upenn.edu} \\}

\usepackage{amsmath,amssymb}
\usepackage{amsthm}
\usepackage{stmaryrd}
\usepackage{algorithm,algpseudocode}
\usepackage{graphicx}
\usepackage{booktabs}
\usepackage{enumitem}

\newtheorem{definition}{Definition}[section]
\newcommand{\para}[1]{\vspace{4pt}\noindent\textbf{#1}}

\begin{document}
\maketitle
\begin{abstract}
A key challenge facing natural language interfaces is enabling users to understand the capabilities of the underlying system. We propose a novel approach for generating explanations of a natural language interface based on semantic parsing. We focus on counterfactual explanations, which are post-hoc explanations that describe to the user how they could have minimally modified their utterance to achieve their desired goal. In particular, the user provides an utterance along with a demonstration of their desired goal; then, our algorithm synthesizes a paraphrase of their utterance that is guaranteed to achieve their goal. In two user studies, we demonstrate that our approach substantially improves user performance, and that it generates explanations that more closely match the user's intent compared to two ablations.\footnote{Code available at: \url{https://github.com/georgeto20/counterfactual_explanations}.}
\end{abstract}

\section{Introduction}

Semantic parsing is a promising technique for enabling natural language user interfaces~\cite{ge2005statistical,artzi2013weakly,berant2013semantic,wang2015building}. However, a key challenge facing semantic parsing is the richness of human language, which can often encode concepts (e.g., ``circle'') that do not exist in the underlying system or are encoded using different language (e.g., ``ball''). Thus, human users can have trouble providing complex compositional commands in the form of natural language to such systems.

One approach to addressing this issue is to develop increasingly powerful models for understanding natural language~\cite{gardner2018neural,yin2018tranx}. While there has been enormous progress in this direction, there remains a wide gap between what these models are capable of compared to human understanding~\cite{lake2018generalization}, manifesting in the fact that these models can fail in unexpected ways~\cite{ribeiro2016should}. This gap can be particularly problematic for end users who do not understand the limitations of machine learning models, since it encourages the human user to provide complex commands, but then performs unreliably on such commands.

Thus, an important problem is to devise techniques for explaining these models. Generally speaking, a range of techniques have recently been developed for explaining machine learning models. The first technique is to use models that are intrinsically explainable, such as linear regression or decision trees. However, in the case of semantic parsing, such models may achieve suboptimal performance, and furthermore it is not clear that the structure of these models would be useful to end users. A second technique is to train a blackbox model, and then approximate it using an interpretable model. Then, the interpretable model can be shown to the human user to explain the high-level decision-making process underlying the blackbox model. However, this approach also suffers from the fact that showing a decision tree or regression model is likely not useful to an end user.

Instead, we consider an alternative form of explanation called a \emph{counterfactual explanation}~\cite{wachter2017counterfactual}. These explanations are designed to describe alternative outcomes to the user. In particular, given a prediction for a specific input, they tell the user how they could have minimally modified that input to achieve a different outcome. As an example, suppose a bank is using a machine learning model to help decide whether to provide a loan to an individual; if that individual is denied the loan, then the bank can provide them with a counterfactual explanation describing how they could change their covariates (e.g., increase their income) to qualify for a loan.

We propose a novel algorithm for computing counterfactual explanations for semantic parsers. In particular, suppose that a user provides a command in the form of a natural language utterance. If the natural language interface fails to provide the desired result, then our goal is to explain how the user could have modified their utterance to achieve the desired result. To this end, we have the human additionally provide the desired result. Then, we compute an alternative utterance that the semantic parser correctly processes while being as similar as possible to the original utterance. Intuitively, this explanation enables the user to modify their language to reliably achieve their goals in future interactions with the system.

We evaluate our approach on the BabyAI environment~\cite{chevalier2018babyai}, where the human can provide a virtual agent with commands to achieve complex tasks such as ``pick up the green ball and place it next to the blue box''. We perform two user studies, which demonstrate that our approach both produces correct explanations (i.e., match the user's desired intent), and that it substantially improves the user's ability to provide valid commands.

\begin{figure}
\centering
\includegraphics[width=0.25\textwidth]{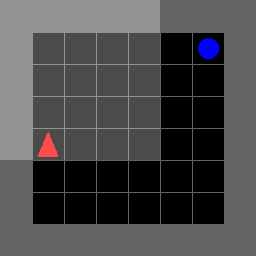} \\
\textbf{User command 1:} ``Go to the blue circle'' \\
\textbf{User command 2:} ``Go to the top right'' \\
\textbf{Our explanation:} ``Go to the blue ball''
\caption{Example BabyAI task (from~\citet{chevalier2018babyai}), utterances, and our explanation.}
\label{fig:example}
\end{figure}

\para{Example.}
In Figure~\ref{fig:example}, we show an example of a BabyAI task along with a user-provided utterance commanding the agent to go to the blue ball. The first command corresponds to a valid program, but cannot be understood by the semantic parser due to the use of the terminology ``circle'' instead of ``ball''. The second command uses the construct ``top right'' that does not exist in the language. In both cases, the user provides a demonstration where the agent navigates next to the blue ball, upon which our approach generates the explanation shown.

\para{Related work.}
There has been a great deal of recent interest in providing explanations of blackbox machine learning models, focusing on explaining why the model makes an individual prediction~\cite{ribeiro2016should,lei2016rationalizing,ribeiro2018anchors,alvarez2018towards,liu2018towards}, or achieving better understanding of the limitations of models~\cite{wallace2019universal,ribeiro2020beyond}. In contrast, our goal is to explain how the input can be changed to achieve a desired outcome, which is called a counterfactual explanation~\cite{wachter2017counterfactual,ustun2019actionable}. There has been interest in 
improving the performance of semantic parsers through interaction~\cite{wang2016learning,wang2017naturalizing}; our approach is complementary to this line of work, since it aims to make the system more transparent to the user. There has also been work on leveraging natural language descriptions to help generate counterfactual explanations for image classifiers~\cite{hendricks2018generating}, but not tailored at counterfactual predictions for natural language tasks; specifically, while their approach produces counterfactual explanations in natural language, they are for image predictions rather than text predictions.

For natural language processing tasks, a key challenge is that the input space is discrete (e.g., a natural language utterance); for such settings, there has been work on algorithms for searching over combinatorial spaces of counterfactual explanations~\cite{ross2021explaining,wu2021polyjuice,ross2021learning}. However, even for these approaches, the output space is typically small (e.g., a binary sentiment label). In contrast, semantic parsing has highly structured outputs (i.e., programs), requiring significantly different search procedures to find an explanation that produces the correct output. To address this challenge, we define a search space over counterfactual explanations for semantic parsing such that search is tractable.

\section{Algorithm}

\para{Problem formulation.}
We consider the problem of computing counterfactual explanations for a semantic parsing model $f_{\theta}:\Sigma^*\to\Pi$. In particular, we assume the user provides a command in the form of an utterance $s\in\Sigma^*$, with the goal of obtaining some denotation $y\in\mathcal{Y}$. To achieve the user's goal, the semantic parsing model produces a program $\pi=f_{\theta}(s)\in\Pi$, and then executes the program to obtain denotation $y=\llbracket\pi\rrbracket\in\mathcal{Y}$, where $\llbracket\cdot\rrbracket:\Pi\to\mathcal{Y}$ (called the \emph{semantics} of $\Pi$) maps programs to outputs.

In this context, our goal is to provide explanations to the user to help them understand what utterances can be correctly understood and executed by the underlying system. In particular, we assume the user has provided an utterance $s_0$, but the output $\llbracket f_{\theta}(s_0)\rrbracket$ is not the one that they desired. Then, we ask the user to provide their desired output, after which we provide them with an alternative utterance $s^*$ that is semantically similar to $s_0$ but successfully achieves $y_0$. Formally:
\begin{definition}
\rm
Given an utterance $s_0\in\Sigma^*$ and a desired output $y_0\in\mathcal{Y}$, the \emph{counterfactual explanation} for $s_0$ and $y_0$ is the sentence
\begin{align*}
s^*=&\operatorname*{\arg\min}_{s\in L}d(s,s_0)
~~\text{subj. to}~~\llbracket f_{\theta}(s)\rrbracket=y_0,
\end{align*}
where $d$ is a semantic similarity metric and $L\subseteq\Sigma^*$ is the search space of possible explanations.
\end{definition}
The goal is that examining $s^*$ should help the user provide utterances that are more likely to be correctly processed in future interactions.

\para{Search space of explanations.}
A key challenge in generating natural language expressions is how to generate expressions that appear natural to the human user. To ensure that our explanations are natural, we restrict to sentences generated by a context-free grammar (CFG) $C$. In particular, we consider explanations in the form of sentences $s\in L(C)\subseteq\Sigma^*$ (where $\Sigma$ is the vocabulary and $L(C)$ is the language generated by $C$). We restrict to sentences with parse trees of bounded depth $d$ in $C$; we denote this subset by $L_d(C)$.
In addition, we assume sentences $s\in L_d(C)$ are included in the dataset used to train the semantic parser $f_{\theta}$ to ensure it correctly parses these sentences.

\para{Semantic similarity.}
Our goal is to compute a sentence $s\in L_d(C)$ that is semantically similar to the user-provided utterance $s_0$. To capture this notion of semantic similarity, we use a pretrained language model $x=g_{\theta}(s)$ that maps a given sentence $s$ to a vector embedding $x\in\mathbb{R}^k$. Then, we use cosine similarity in this embedding space to measure semantic similarity. In particular, we use the distance
$d(s,s_0)=1-\text{sim}(g_{\theta}(s),g_{\theta}(s_0))$,
where $\text{sim}(x,x')$ is the cosine similarity.

\para{Goal constraint.}
Finally, we want to ensure that the provided explanation successfully evaluates to the user's desired denotation $y_0$. For a given utterance $s$, we can check this constraint simply by evaluating $y=\llbracket f_{\theta}(s)\rrbracket$ and checking if $y=y_0$.

\begin{algorithm}[t]
\begin{algorithmic}
\Procedure{Explain}{$s_0,y_0$}
\State $(s^*,c^*)\gets(\varnothing,-\infty)$
\For{$s\in L_d(C)$}
\If{$\llbracket f_{\theta}(s)\rrbracket=y_0$}
\State $c\gets\text{sim}(g_{\theta}(s),g_{\theta}(s_0))$
\State \textbf{if} $c>c^*$ \textbf{then} $s^*,c^*\gets s,c$ \textbf{end if}
\EndIf
\EndFor
\State \Return $s^*$
\EndProcedure
\end{algorithmic}
\caption{Our algorithm for computing counterfactual explanations for a semantic parser $f_{\theta}$.}
\label{alg:main}
\end{algorithm}

\para{Overall algorithm.}
Given user-provided utterance $s_0$ and desired denotation $y_0$, the counterfactual explanation problem is equivalent to:
\begin{align*}
s^*=
&\operatorname*{\arg\max}_{s\in L_d(C)}\text{sim}(g_{\theta}(s),g_{\theta}(s_0)) \\
&\text{subj. to}~\llbracket f_{\theta}(s)\rrbracket=y_0.
\end{align*}
Assuming $L_d(C)$ is sufficiently small, we can solve this problem by enumerating through the possible choices $s\in L_d(C)$ and choosing the highest scoring one that satisfies the constraint. In practice, we may be able to exploit the structure of the constraint to prune the search space. Our approach is summarized in Algorithm~\ref{alg:main}.

\section{Experiments}

We perform two user studies to demonstrate
(i) correctness: our explanations preserve the user's original intent, and (ii) usefulness: our explanations improve user performance.

\subsection{BabyAI Task}

We evaluate our approach on BabyAI~\cite{chevalier2018babyai} adapted to our setting. In this task, the human can provide commands to an agent navigating a maze of rooms containing keys, boxes, and balls. The goal is defined by the combination of the agent position and the environment state (e.g., the agent may need to place a ball next to a box).
Atomic commands (e.g., going to, picking up, or putting down an object) can then be composed in sequence to achieve complex goals. In our setup, $s_0$ is a natural language command, and $y_0$ is a demonstration in the form of a trajectory the agent could take to achieve the desired goal.

This task comes with a context-free grammar of natural language commands, which we use as the space of possible explanations. Next, we train a semantic parser to understand commands from this grammar. Since utterances in this grammar correspond one-to-one with programs, we can generate training data. We generate 1000 training examples $(s,\pi)$ consisting of an utterance $s$ along with a program $\pi$, and train TranX~\cite{yin2018tranx} to predict $\pi=f_{\theta}(s)$. For semantic similarity, we use a pretrained DistilBERT model $g_{\theta}$~\cite{devlin2018bert,sanh2019distilbert} to embed utterances $s$.

Handling the goal constraint is more challenging, since the denotation can be nondeterministic---in particular, multiple different trajectories can be used to achieve a single goal (e.g., there are multiple paths the agent can take to a given object). Thus, if we na\"{i}vely take the denotation of a program to be a single trajectory that achieves the goal, then this trajectory may be different than the given demonstration even if the demonstration also achieves the goal. To address this issue, we instead enumerate the set $\Pi_y$ of all possible programs that are consistent with the given demonstration $y$, up to a bounded depth (selected so that $\Pi_y$ is large enough while ensuring that the experiments still run quickly). Then, we replace the constraint $\llbracket f_{\theta}(s)\rrbracket=y_0$ with a constraint saying that $f_{\theta}(s)$ is in this set---i.e.,
$f_{\theta}(s)\in\Pi_{y_0}$.

\subsection{Correctness of Explanations}
\label{sec:correct}

We evaluate whether our explanations are valid paraphrases of the user's original command.

\para{Baselines.}
We compare to two ablations of our algorithm. The first one omits the goal constraint $f_{\theta}(s)\in\Pi_y$; thus, it simply returns the explanation that is most semantically similar to the user-provided utterance $s_0$. Intuitively, this ablation evaluates the usefulness of the goal constraint.

The second ablation ignores $s_0$, and returns an explanation $s$ such that $f_{\theta}(s)\in\Pi_{y_0}$; we choose $s$ to minimize perplexity according to GPT-2. Intuitively, this ablation measures the usefulness of specializing the explanation to the user's utterance.

\para{Setup.}
We selected 17 BabyAI tasks by randomly sampling BabyAI levels until we obtain a set of tasks of varying difficulty. For example, Task 1 has the simple goal ``go to the green ball'', while Task 10 has the more complex goal ``pick up a green key, then put the yellow box next to the grey ball''.

Then, our experiment proceeds in two phases. In the first phase, we use Amazon Mechanical Turk (AMT) to collect natural language commands for the agent. For each of our 17 tasks, we show the user a video of the BabyAI agent achieving the task, and then ask them to provide a single command that encodes the goal.
In total, we obtained 127 commands (one per user) for each task. Next, for each user instruction, we find the counterfactual explanation according to our algorithm and the two ablations described above.

In the second phase, we conduct a second AMT study to evaluate the correctness of these explanations. In particular, for each of our 17 tasks, we show each participant a single command for that task (chosen randomly from the 127 commands in the first phase), along with the three generated explanations and the video of the agent achieving that task. Then, we ask the user to choose the explanation that is closest in meaning to the original command. We obtained 50 responses.

\para{Results.}
In Table~\ref{tab:results}, we show the fraction of times users in the second phase selected each explanation, averaged across both users and tasks. Our approach significantly outperforms GPT-2, which is unsurprising since this ablation makes no effort to preserve the user's intent. Our approach also outperforms the ablation without the goal constraint, demonstrating the usefulness of this constraint.

\begin{table}
\centering\small
\begin{tabular}{lrr}
\toprule
\multicolumn{1}{c}{\textbf{Approach}} &
\multicolumn{1}{c}{\textbf{Correctness}} &
\multicolumn{1}{c}{\textbf{Usefulness}} \\
\midrule
Ours & 41.4 $\pm$ 1.48\% & 50.8 $\pm$ 2.22\% \\
No demo & 34.0 $\pm$ 1.42\% & 49.2 $\pm$ 2.01\% \\
GPT-2 & 24.6 $\pm$ 1.29\% & 46.2 $\pm$ 1.96\% \\
No training & \multicolumn{1}{c}{--} & 10.2 $\pm$ 0.74\% \\
\bottomrule
\end{tabular}
\caption{Correctness: The frequency at which users chose the explanation generated using the corresponding approach as the best match. Usefulness: The percentage of user utterances correctly parsed (averaged across the last 10 tasks), where users are given explanations generated by the corresponding approach.}
\label{tab:results}
\end{table}

\subsection{Usefulness of Explanations}

Next, we evaluate whether providing explanations can make it easier for users to provide commands that can be understood by our semantic parser.

\para{Baselines.}
In addition to the two ablations in Section~\ref{sec:correct}, we also compare to a baseline where the user is not provided with any explanation.

\para{Setup.}
We run an AMT study similar to the first phase of our study in Section~\ref{sec:correct}, except immediately after providing a command for a task, each user is shown an explanation for their command and that task. We collected 50 user responses.

\para{Results.}
For each user command $s_0$, we run our semantic parser to obtain the corresponding program and check whether it is in the set of programs valid for that task---i.e., whether $f_{\theta}(s_0)\in\Pi_{y_0}$. 
Table~\ref{tab:results} shows the success rate across all users and the last 10 tasks; we restrict to the last 10 to give the user time to learn to improve their performance.
Users not provided any explanations performed very poorly overall. The remaining approaches performed similarly; our explanations led to the best performance, followed closely by the ablation without the demonstration, with a wider gap to the ablation that ignores the user utterance. Thus, personalizing the explanation to the user based on their utterance helps improve performance.




\section{Conclusion}

We have proposed a technique for explaining how users can adapt their utterances to interact with a natural language interface. Our experiments demonstrate how our explanations can be used to significantly improve the usability of semantic parsers when they are limited in terms of their semantic understanding. While any explanations are already very useful, we show that personalizing explanations can further improve performance.

A key design choice in our approach is to construct a synthetic grammar from which counterfactual explanations are generated. In a realistic application, the semantic parsing model can be trained on a combination of synthetic data and real-world data, enabling our approach to be used in conjunction with the synthetic grammar. A key direction for future work is extending our approach to settings where such a grammar is not available. In our experience, a key challenge in this setting is that the generated text can be unnatural, possibly due to the constraints imposed on the search space.

\section*{Acknowledgments}

We gratefully acknowledge support from NSF CCF-1917852 and CCF-1910769. The views expressed are those of the authors and do not reflect the official policy or position of the U.S. Government.


\bibliography{paper}
\bibliographystyle{acl_natbib}


\end{document}